\documentclass[runningheads]{llncs}

\usepackage{tikz}
\usepackage{float}
\usepackage{comment}
\usepackage{amsmath,amssymb} 
\usepackage{color}
\usepackage{multirow}
\usepackage{eqparbox}
\usepackage{arydshln}
\usepackage{graphicx}
\usepackage{algorithm}
\usepackage{algorithmic} 
\usepackage{capt-of}
\usepackage{subfigure}
\usepackage{verbatim}
\usepackage{mathrsfs} 
\usepackage{hyperref}
\usepackage{makecell}
\usepackage{booktabs}
\usepackage[accsupp]{axessibility}  

\newcommand{\xin}[1]{\textcolor{red}{[Xin]: #1}}

\begin{document}
\pagestyle{headings}
\mainmatter

\title{FedVLN: Privacy-preserving Federated Vision-and-Language Navigation} 



\titlerunning{FedVLN}
%
\author{Kaiwen Zhou \and
Xin Eric Wang}
\authorrunning{K. Zhou and X. E. Wang}
%
\institute{University of California, Santa Cruz, CA 95064, USA\\
\email{kzhou35@ucsc.edu,xwang366@ucsc.edu}}

\maketitle

\begin{abstract}
Data privacy is a central problem for embodied agents that can perceive the environment, communicate with humans, and act in the real world.
While helping humans complete tasks, the agent may observe and process sensitive information of users, such as house environments, human activities, etc.
In this work, we introduce privacy-preserving embodied agent learning for the Vision-and-Language Navigation (VLN) task, where an embodied agent navigates house environments by following natural language instructions. 
We view each house environment as a local client, which shares nothing other than local updates with the cloud server and other clients, and propose a novel Federated Vision-and-Language Navigation (FedVLN) framework to protect data privacy during both training and pre-exploration.
Particularly, we propose a decentralized federated training strategy to limit the data of each client to its local model training and a federated pre-exploration method to do partial model aggregation to improve model generalizability to unseen environments.
Extensive results on R2R and RxR datasets show that, decentralized federated training achieve comparable results with centralized training while protecting seen environment privacy, and federated pre-exploration significantly outperforms centralized pre-exploration while preserving unseen environment privacy. Code is available at \href{https://github.com/eric-ai-lab/FedVLN}{https://github.com/eric-ai-lab/FedVLN}.

\keywords{Privacy-preserving Embodied AI, Vision-and-Language Navigation, Federated Learning}
\end{abstract}

\section{Introduction}

\begin{figure}[t]
\centering
\subfigure[Centralized VLN learning]{
\begin{minipage}[t]{0.45\textwidth}
\centering
\includegraphics[width=5.5cm]{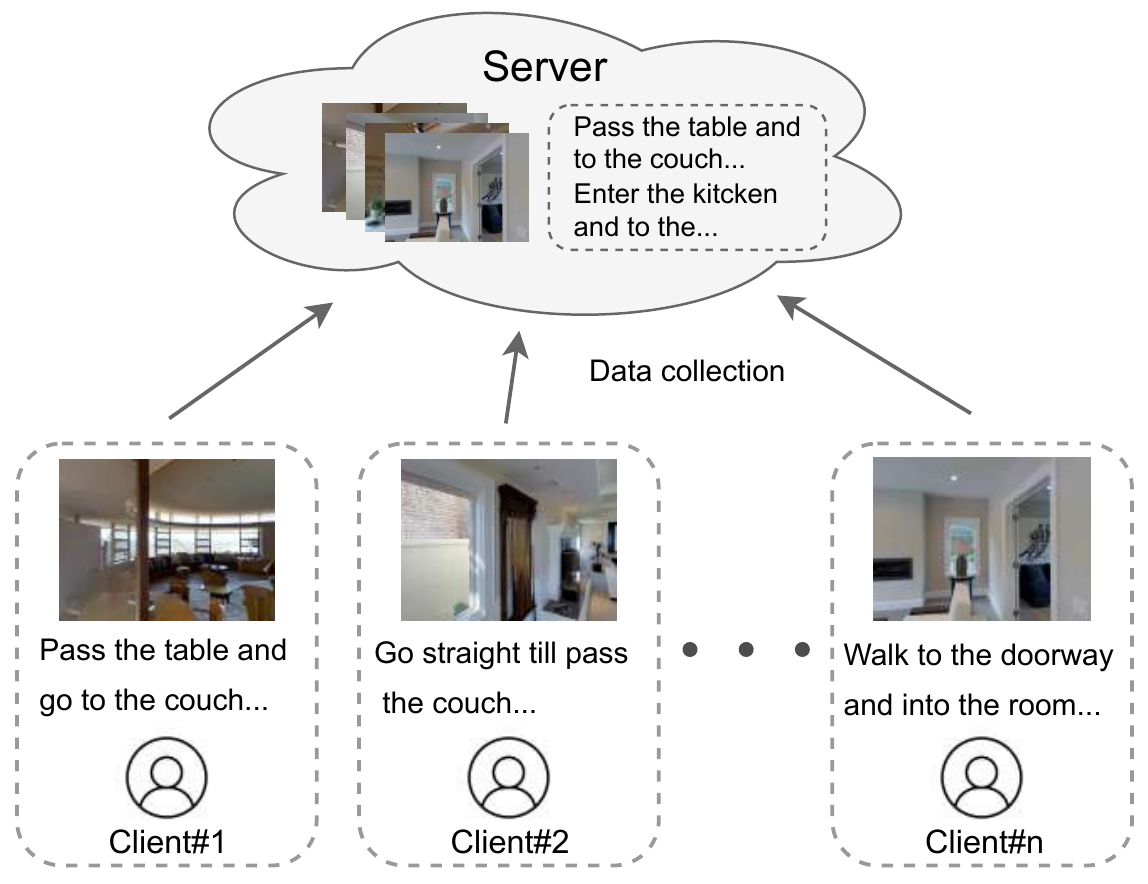}
\end{minipage}
\label{centralized vln}
}
\subfigure[Federated VLN learning]{
\begin{minipage}[t]{0.45\textwidth}
\centering
\includegraphics[width=5.5cm]{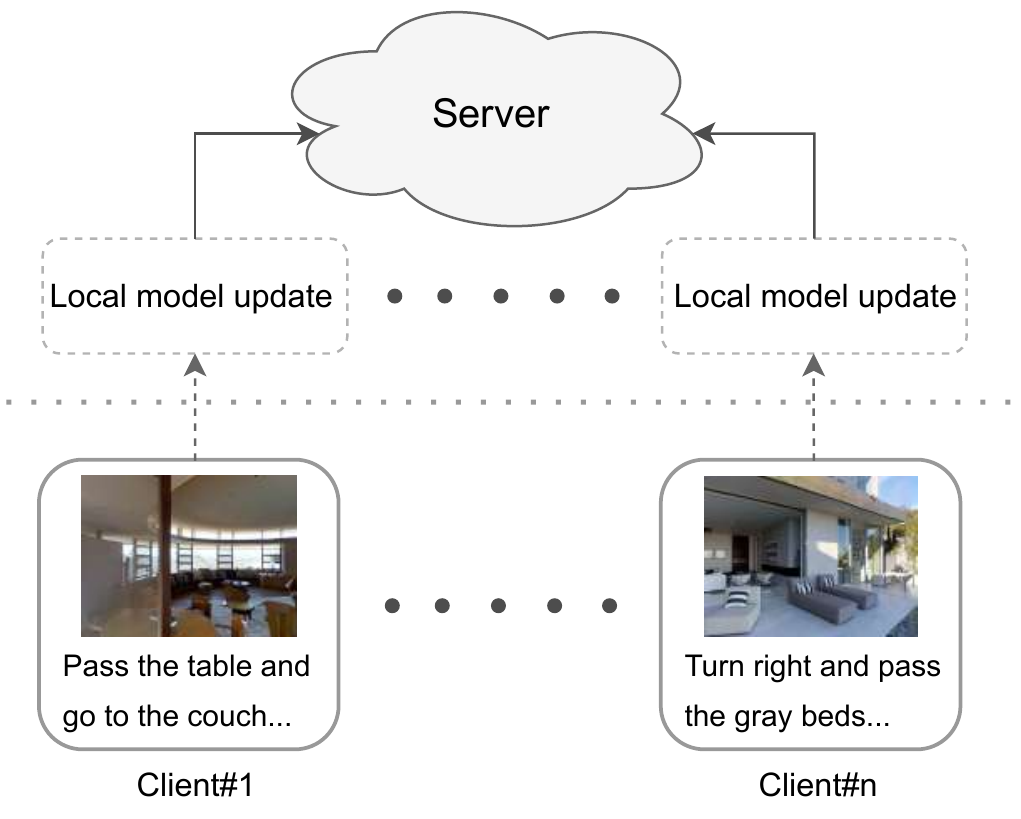}
\end{minipage}
\label{fig:fedvln}
}
\caption{Data privacy: centralized VLN learning \textit{vs.} our federated VLN learning. Existing VLN approaches centralize all the client data in a server, including house environments and user instructions, which ignores users' privacy concerns. Our federated VLN framework keeps client data used only locally and the server receives nothing other than local model updates, so the client data privacy is preserved.} 
\label{centalized vs fed}
\end{figure}

A long-term goal of AI research is to build embodied agents that can perceive the environment, communicate with humans, and perform real-world tasks to benefit human society. 
However, since the agent interacts closely with humans and environments, it often receives sensitive information during training and inference. 
For example, as shown in Fig.~\ref{centralized vln}, in the task of Vision-and-Language Navigation (VLN)~\cite{r2r}, where an agent learns to navigate towards a target location in an indoor environment given natural language instruction, the training and inference data may include private information such as what the user's house looks like, what the user said, and what the user did. Data privacy is a central problem for building trustworthy embodied agents but seldomly studied before~\cite{gu-etal-2022-vision}. Thus, in this work, we introduce privacy-preserving embodied agent learning for the task of vision-and-language navigation.

VLN models are typically trained on seen environments with ground-truth instruction-trajectory pairs and then deployed to unseen environments without any labeled data. 
After deployment, the agent may explore the unseen environment and adapt to the new environment for better performance, which is known as pre-exploration. 
However, as shown in Fig.~\ref{centralized vln}, most of the existing methods assemble all the data in a server to train a navigation agent for both seen environment training and unseen environment pre-exploration.
This is not practical in reality since users may not want to share the data in their own house due to privacy concerns. 
Privacy-preserving VLN requires the agent to protect the data privacy during both seen environment training and unseen environment pre-exploration while maintaining comparable navigation performance.



In this paper, we propose a novel Federated Vision-and-Language Navigation (FedVLN) framework, to address the aforementioned data privacy issues and improve the adaptation performance on unseen environments at the same time. 
Specifically, on the seen environment training stage, as shown in Fig.~\ref{fig:fedvln}, we treat each house environment as a client. The client's local models (a VLN agent for navigation and a speaker for data augmentation) are trained on private local data, and then the model updates are sent to the server for model aggregation. No private data except the local model updates will be shared with the server and there is no communication between clients.
During pre-exploration, we train the client models on seen environments and unseen environments simultaneously under federated learning paradigm---the client models do partial model aggregation (language encoder only) and partial local adaptation, enabling better adaptation to the local visual environment while maintaining general language understanding. 
Under our FedVLN framework, users do not need to share their data with any other party, thus the privacy of training data and inference data is protected.

Our experiments on Room-to-Room (R2R)~\cite{r2r} and Room-across-Room (RxR) \cite{rxr}\footnote{We conduct experiments on the English data of the RxR dataset.} datasets validate the effectiveness of our framework. Our federated learning framework achieves comparable results with centralized training while preserving data privacy. More importantly, on the pre-exploration stage, we show that centralized pre-exploration hinders the agent from adapting to each specific environment, and our federated pre-exploration method achieves the best performance among prior pre-exploration methods such as centralized~\cite{RCM,Envdrop} and environment-based pre-exploration~\cite{APS}. Our contributions are three-fold:
\begin{itemize}
  \item We are the first to discuss data privacy concerns for vision-and-language navigation and define the privacy-preserving embodied AI problem for the two learning stages in VLN. 
  \item We propose a novel federated learning framework for privacy-preserving VLN to ensure that users do not need to share their data to any party.
  \item Extensive results on R2R and RxR show that our federated learning framework not only achieves comparable results with centralized training, but also outperforms centralized and environment-based pre-exploration methods.
\end{itemize}

\section{Related Work}
\subsection{Vision-and-Language Navigation}
With the development of deep learning and human's vision of more helpful AI agents, embodied AI becomes an emerging research area. Vision-and-language navigation(VLN)~\cite{r2r,rxr,jain-etal-2019-room-for-room,reverie,zhu-etal-2022-diagnosing} is one of the most popular tasks of embodied AI, in which an embodied agent learns to navigation to a goal location in indoor environments following language instruction and given dynamic visual information. Anderson et al. \cite{r2r} first propose a LSTM-based seq-to-seq model for navigation. For better understanding vision-and-language information, there are works working on vision-and-language pre-training~\cite{Hao_2020_CVPR_prevalent,oscar,rec-vlnbert,Qi_2021_ICCV_object,Guhur_2021_ICCV_airbert} and model structures~\cite{rec-vlnbert,chen2021hamt,Gao_2021_CVPR_room-and-Object,xiang-etal-2020-learning}. Reinforcement learning and navigation planning methods were also introduced into VLN to perform better action decisions~\cite{RCM,Wang_2018_ECCV_look_before_you_leap,Krantz_2021_ICCV_waypoint,chasing_ghosts}. Limited labeled data was another bottleneck to train a better model. To this end, Fried et al.~\cite{Fried2018SpeakerFollowerMF} propose a speaker-follower model which can generate pseudo instructions for a sampled path by a trained speaker. Further, to mitigate the gap between seen and unseen environments, pre-exploration was proposed~\cite{RCM,Envdrop,APS}, which can learn and adapt to new environments after deployment. 
However, most current research ignores the practicality in real-life application scenarios, especially data privacy issues. Fu et al.~\cite{APS} consider the implementation problem of pre-exploration and proposed environment-based pre-exploration, but they did not consider the privacy issue of training data. Also, we showed that environment-based pre-exploration might suffer from data scarcity and data bias. 


\subsection{Privacy-preserving Machine Learning}
Over the years, researchers propose many methods~\cite{FasterCrypto,PATE,fedavg,texthide} to address different data privacy problems~\cite{ModelInversion,MembershipInfernece,PropertyInference,MembershipInferenceMT} in machine learning. 
First, during the training stage, if the training data are from different parties, sharing their data with other parties might leads to privacy concerns. At the inference stage, there are multiple privacy attacks, especially in the scenario of Machine Learning as a Service (MLaaS), in which cloud providers offer machine inference hosted on the cloud~\cite{FasterCrypto}. For example, membership inference attack~\cite{MembershipInfernece} can judge if a specific data sample exists in training data, model inversion attack~\cite{ModelInversion,SecretRevealer} aims to infer training data given white-box or black-box access to the model. Also, in MLaaS, users might not be willing to directly upload their data to the cloud server~\cite{FasterCrypto}. Facing these privacy problems for training data and inference data, researchers propose many privacy-preserving methods, including federated learning, homomorphic encryption, differential privacy, etc~\cite{fedavg,SHE,PATE,Li_2021_CVPR}. 
However, most of their work focuses on single modality tasks and static data, and seldomly study the data privacy of embodied AI. In embodied AI tasks like vision-and-language navigation, the data contains more human-robot interaction and more complex private information, such as corresponding language-image pairs, dynamic visual information in the indoor environments. VLN also has a unique training stage, pre-exploration. Both of these may make the privacy problems and solutions for VLN more complex. In our work, we elaborate on privacy-preserving VLN training scenarios and propose a solution.

\subsection{Federated Learning}
Federated learning~\cite{fedavg} is a technique that allows client models to train locally and then be sent to the central server for model aggregation. In this way, the clients do not need to send their sensitive data to any party. Thus the privacy of training data is protected. The first federated learning algorithm~\cite{fedavg} uses weighted sum for aggregating clients' models. Later, researchers proposed different federated learning algorithms for heterogeneous data distribution and personalization~\cite{Li_2021_CVPR,ShareRep,HsuECCV2020Real-World,Personalized_Cross-Silo}. Especially, Collins et al.~\cite{ShareRep} proposed to keep classification head locally for personalization. Compared with our framework, they were trying to solve the problem of label heterogeneity and learn a general data representation, and their setting does not have the difference between validation data and training data. Reddi et al.~\cite{reddi2021adaptive} summarized these first-order aggregation methods into one framework as {\large F}ED{\large O}PT, whose server aggregation is:
\begin{equation}
\begin{aligned}
     w_{t+1} = {\rm {\large S}ERVER{\large O}PT}(w_{t}, -\Delta w_{t}, \eta, t)
\end{aligned}
\end{equation}
Where SERVEROPT is the aggregation algorithm, $\eta$ is server learning rate.

Application wise, federated learning framework has been used on various tasks in computer vision~\cite{Guo_2021_CVPR_Multi-Institutional,HsuECCV2020Real-World} and natural language processing~\cite{lu2021federated,huang-etal-2020-federated}. Recently, there are also some works for federated learning on multi-modal machine learning~\cite{Liu_Wu_Ge_Fan_Zou_2020_fed_ground,zhao2022multimodal}. Zhao, et al.~\cite{zhao2022multimodal} try horizontal federated learning(FL), vertical FL, and Federated Transfer Learning on different multi-modal tasks and datasets, and~\cite{Liu_Wu_Ge_Fan_Zou_2020_fed_ground} using semi-supervised FL to extract hidden representations of multi-modality. However, the tasks they discussed is not embodied agent for individual users. In vision-and-language navigation, the training paradigm is different from formerly discussed tasks, which has two different training objectives, training scenarios in two training stages. To solve this, we proposed a novel Federated Vision-and-Language Navigation(FedVLN) framework.

\section{Privacy-preserving Vision-and-Language Navigation}
\subsection{Vision-and-Language Navigation (VLN)} \label{background}
The goal of the VLN task is to navigate from a given location and reach a destination following natural language instruction. The task can be formally defined as follow: given an language instruction $U = \{u_{1}, u_{2}, ..., u_{n}\}$. At each step, the agent will receive current visual information $v_{t}$ as input. The agent will need to choose an action $a_{t}$ at each step based on the instruction $U$, current/history visual information$\{v_{\tau}\}_{\tau=1}^{t}$, and history actions$\{a_{\tau}\}_{\tau=1}^{t-1}$. The agent's state, which consists of the agent's navigation history and current spatial location, will change according to the agent's action. The navigation terminates when the agent chooses a `stop' action. The environments that contain labeled training data are seen environments. There are also unseen environments that do not have training data and are invisible during training. 

\noindent\textbf{VLN agents~} In general, VLN agents consist of a language encoding module to understand the instruction, a trajectory encoder to encode visual observation and actions, 
and a multimodal decision module to jointly process multi-modal information including encoded language information $L_{enc}$, visual information $V_{enc}$, and action information $A_{enc}$ and predict the next action $a_{t}$: 
\begin{gather}
    L_{enc} = E_{L}(u_{1}, u_{2},...,u_{n})\\
    V_{enc}, A_{enc} = E_{T}(v_{1}, v_{2},...,v_{t}, a_{1}, a_{2},...,a_{t-1}) \label{trajectory encoder}\\
    a_{t} = M(L_{enc}, V_{enc}, A_{enc})
\end{gather}

\noindent\textbf{Speaker-based data augmentation~}
To tackle the problem of data scarcity, Fried et al.~\cite{Fried2018SpeakerFollowerMF} propose a back-translation speaker model which can generate corresponding instructions $U$ from the visual information and action sequence of sampled routes in the environment:
\begin{equation}
    U = Speaker(v_{1}, v_{2},...,v_{t}, a_{1}, a_{2},...,a_{t})
\end{equation}
The speaker is trained by original labeled route-instruction pairs, which takes the visual and actions information of routes as input and predict the instructions. The generated pseudo instructions along with sampled routes can be the augmented training data for better agent learning. 

\noindent\textbf{Pre-exploration~}
After training on seen environments and deploying on unseen environment, the agent can adapt to the new environment via pre-exploration~\cite{APS,RCM,Envdrop}. There are different variants of pre-exploration includes self-imitation learning~\cite{RCM}, graph-based methods~\cite{spacial_route_prior,Chen_2021_CVPR_topological}, etc. In our work, we consider the paradigm that sampling routes $R^{'}$ from a new environment and generate instructions $I^{'}$ using the trained speaker mentioned before. Then the agent can be trained on the new environment using sampled routes and pseudo instructions$(R^{'},I^{'})$.

\subsection{Privacy-preserving VLN}
\begin{figure}[t]
\centering
\includegraphics[width=9cm]{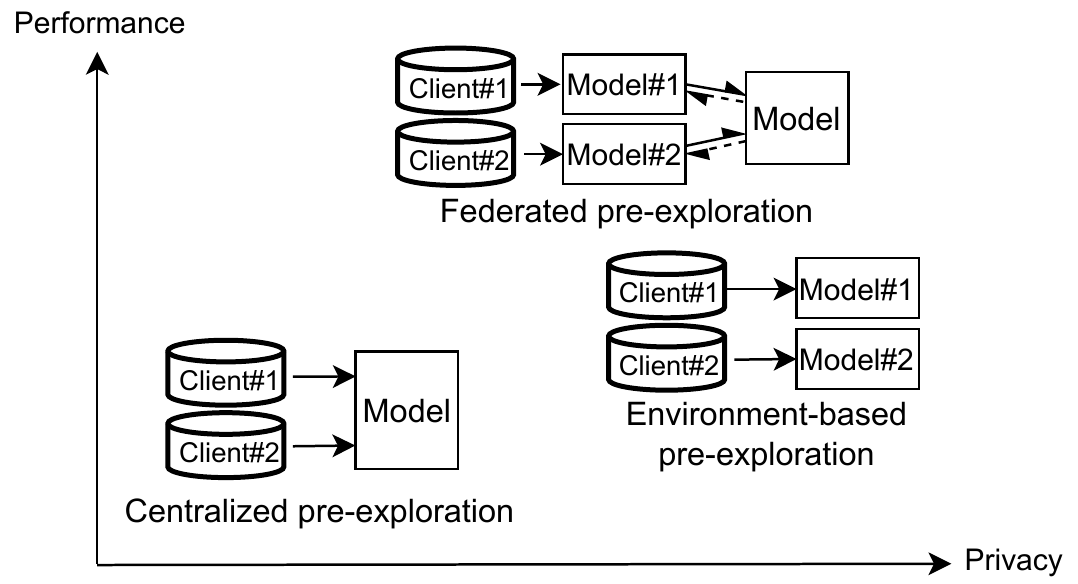}
\caption{Comparison between different pre-exploration strategies on performance-privacy trade-off. Federated pre-exploration achieves the best navigation performance while maintaining good inference data privacy protection.}
\label{private pre explore}
\end{figure}



Considering the data may have sensitive information, users may have different levels of concern about the privacy of their data. In our work, we consider the case that the users do not want their data to be directly shared with the server (e.g., companies) and any other parties. Based on this, we define privacy-preserving vision-and-language navigation learning setting on two training stages: seen environment training and unseen environment pre-exploration. For seen environment training, including the training of navigation agent, speaker model and data augmentation process, no labeled data within the house environment will be directly shared with the server or any other client to prevent the leak of private information. And our primary purpose is to train a model that can generalize well on unseen environments. Thus, we need to utilize all the data indirectly to train one model. 

For pre-exploration, the unlabeled data in unseen environments also can not be shared with others. However, the purpose in this stage is to adapt the model to a specific environment. Thus, training on data in one environment (environment-based pre-exploration) might not be a bad choice. In fact, our experiments show that environment-based pre-exploration performs better than centralized pre-exploration. Nevertheless, as elaborated in Sec.~\ref{fed pre-exploration}, we can indirectly utilize all the data in pre-exploration to boost the performance and preserve privacy. As in Fig.~\ref{private pre explore}, we aim to achieve the best performance-privacy trade-off in pre-exploration. 

\section{The FedVLN Approach}

\begin{figure}[t]
\centering
\subfigure[Decentralized Federated Training]{
\begin{minipage}[t]{0.45\textwidth}
\centering
\includegraphics[width=5.5cm]{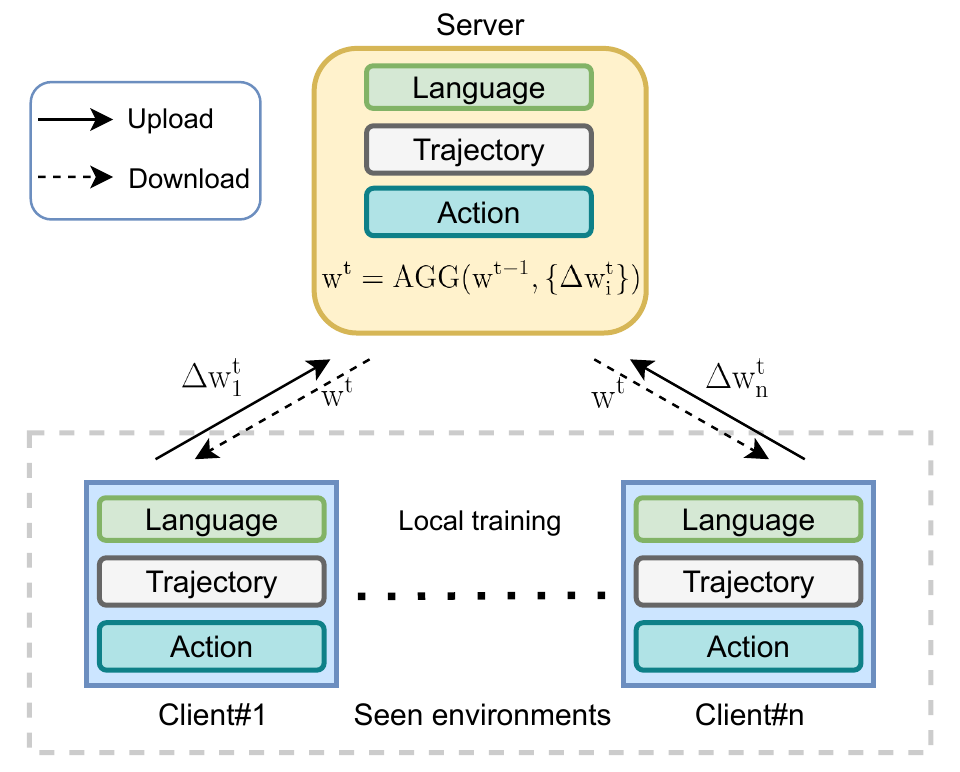}
\label{decentralized fig}
\end{minipage}
}
\subfigure[Federated Pre-exploration]{
\begin{minipage}[t]{0.45\textwidth}
\centering
\includegraphics[width=5.5cm]{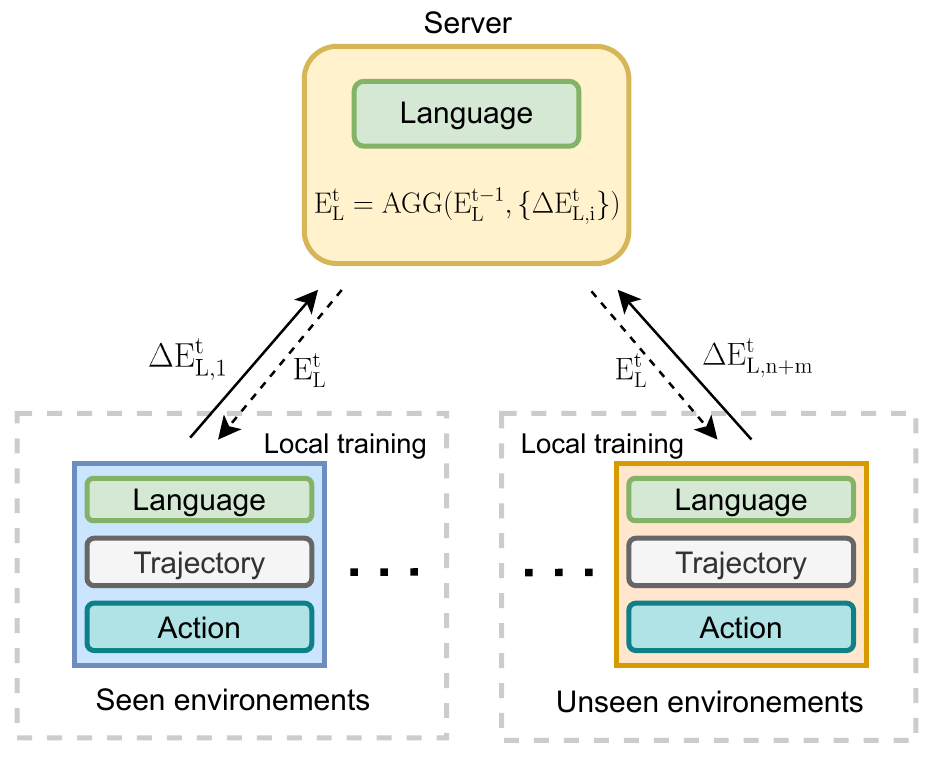}
\label{fed pre-explore fig}
\end{minipage}
\label{fedvln fig}
}

\caption{The FedVLN framework. In the first stage (a), agents in seen environments will be trained on local data and upload the model updates to the server for aggregation (AGG), then download the global model from the server. In the second stage (b), all the agents in seen and unseen environments join the federated training. During local training, all the modules will be optimized, while only the language encoder will be uploaded/downloaded.} 
\label{fedvln fig}
\end{figure}

We propose a federated vision-and-language navigation framework (FedVLN) as shown in Fig.~\ref{fedvln fig}, in which user's data can be kept locally during both training and pre-exploration. In this section, we will introduce our FedVLN framework for two training stages: Decentralized Federated Training and Federated Pre-exploration. In decentralized federated training, each environment has a local agent, which will be trained on local data, then uploaded to the server. Then the global model on the server will be updated by the aggregation of local model updates and sent to all the environments. In federated pre-exploration, to enable the agent to both adapt to the new environment and maintain the ability to understand language, only the language encoder will be shared with the server after local training, instead of sharing the full model. All the agents from seen and unseen environments will join the federated pre-exploration process.  
\subsection{Decentralized Federated Training} 

\noindent\textbf{Original training data~} When training on the original training data, we first divide the VLN dataset by environments. We treat each environment as a client, then assign a local navigation agent $w_{i}^{0}$ on each environment, which is initialized as the same as global navigation agent $w^{0}$. At each communication round between clients and server, a certain percentage of clients will be randomly selected for training, the local agent on each selected client will be trained for a certain number of epochs on their own data $d_{i}$:
\begin{equation}
    w_{i}^{t} = {\rm ClientUpdate}(w^{t-1}, d_{i})
\end{equation}

Where $\rm ClientUpdate$ is the local training process. Then each selected client will send the update $\Delta w_{i, t} = w_{i}^{t} - w^{t-1}$ of their model to the server, and the server will aggregate all the models with a server learning rate $\eta$:
\begin{equation} \label{server aggregation}
    w^{t} = w^{t-1} + \eta \sum_{i\in \phi_{t}}\frac{n_{j}}{\sum_{j\in \phi_{t}}n_{j}} \Delta w_{i}^{t}
\end{equation}
Here the weight of each local model $\frac{n_{j}}{\sum_{j\in \phi_{t}}n_{j}}$ is the proportion of the use's sample in the total training sample of this communication round. 

\noindent\textbf{Augmentation~} 
For data augmentation, we will assign each client a local speaker. Following the federated learning paradigm mentioned above and the training procedure of speaker from Sec.~\ref{background}, at each communication round, each speaker from the selected clients will be trained on the labeled route-instruction pairs in its environment. 

The best global model (according to BLUE score) during the training process will be sent to all clients. Each client can use the speaker model to generate pseudo instructions $I_{i}^{aug}$ for sampled routes within the environment. Then the augmented training of the agent will also follow the federated training process mentioned above, except the local data will be the combination of original data and augmented data $\{(d_{i}, d_{i}^{aug})\}_{i=1}^{n}$. 

Notice that during the whole training process, including original data training, speaker training, and augmented data training, no client share their data to other clients or the server. Thus the training data privacy is preserved.

\begin{figure}[t]
\centering
\begin{minipage}{0.36\textwidth}  
\centering
\resizebox{\columnwidth}{!}{
\begin{tabular}{lrr}
        \toprule
        \textbf{Statistics} &  \textbf{GT} & \textbf{Speaker} \\ 
        \midrule
        Length  & 29.58 & 21.89  \\
        Var(Length) & 155.70 & 20.88 \\
        NoS  & 2.44 & 2.42  \\
        Var(NoS) & 1.21 & 0.47  \\
        \bottomrule
    \end{tabular}
    }
    \captionof{table}{Comparison between ground-truth (GT) and speaker generated instructions on seen validation. NoS means the average number of sentences.}
    \label{instr statistics}
\end{minipage}
\begin{minipage}{0.58\textwidth}
\centering
\begin{minipage}{0.48\textwidth}
\centering
\includegraphics[height = 3.2cm]{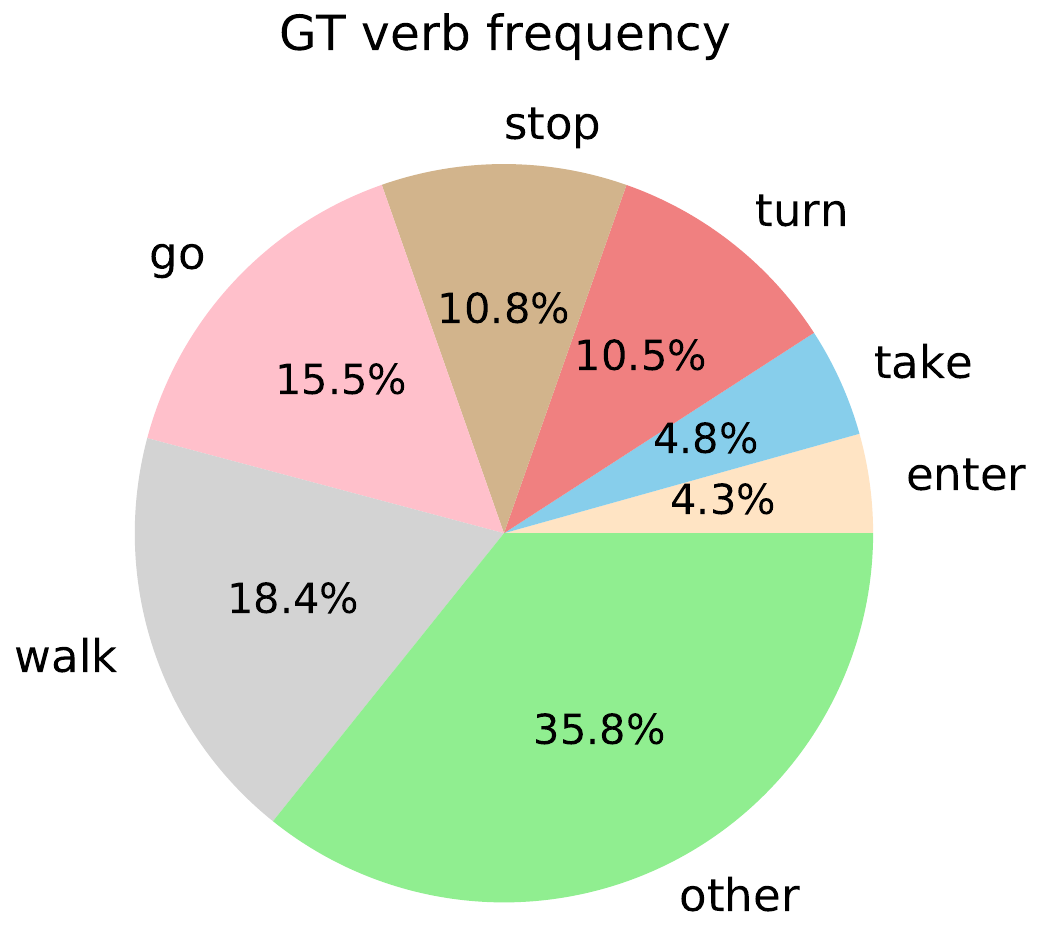}
\end{minipage}
\begin{minipage}{0.48\textwidth}
\centering
\includegraphics[height = 3.3cm]{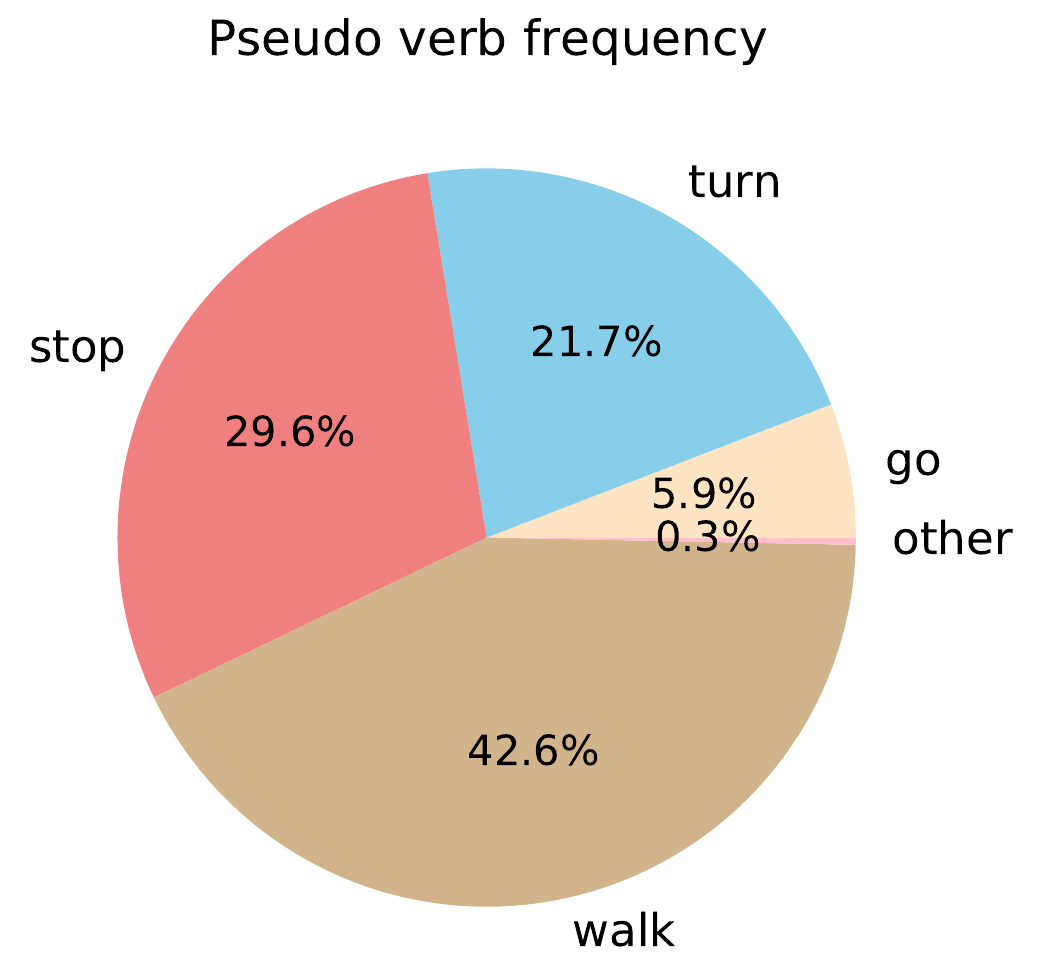}
\end{minipage}
\caption{Comparison of verb frequency between ground truth instructions and generated pseudo instructions. }
\label{verb freq}
\end{minipage}
\end{figure}

\subsection{Federated Pre-exploration~} \label{fed pre-exploration}
Pre-exploration allows the agent to explore the newly deployed environment and update itself based on the new information. From the perspective of data privacy, centralized pre-exploration is impractical here, since it assumes one navigation agent can get access to data from all the unseen environments. Fu et al.\cite{APS} proposed environment-based pre-exploration, which allows each agent to train on only one environment. Thus no data will be shared with other parties, and the data privacy in unseen environments is preserved. From the performance point of view, for centralized training, since the agent is trained on all the data from all the environments, it should have better generalizability. However, training in all the environments may hinder the agent from better adapting to one specific environment. For environment-based pre-exploration, the agent can focus on one specific environment, while the limited data amount and data bias in one environment may lead to a less generalized agent. 

Furthermore, as shown in Table~\ref{instr statistics} and Fig.~\ref{verb freq}, we found that the instructions  generated by the speaker are statistically significantly different from human-generated instructions. Moreover, the language pattern is much simpler than human language. Since current methods only use augmented data with speaker-generated instructions for training during pre-exploration, the agent might suffer from the huge distribution shift between instructions in augmented data and validation data, and can not understand instructions in validation data well. This problem could be even worse on environment-based pre-exploration since the data for one agent is of a smaller amount and from a single environment. 

What is more, according to former research~\cite{diagnosingEnv}, the agent will perform better on seen paths or environments. Thus, the best solution is to maintain the generalizability to understand language and adapt to a specific visual environment. To this end, as in Fig.~\ref{fed pre-explore fig}, we propose federated pre-exploration. In federated pre-exploration, The server will only maintain a global language encoder, which is initialized with the global encoder after decentralized federated VLN training. During each communication round, the server will send the global language encoder $E^{t-1}$ to the selected clients. Then the selected clients will update its language encoder with $E^{t-1}$, and train the full agent on its local data:
\begin{equation}
    E_{L,i}^{t}, E_{T,i}^{t}, M_{i}^{t} = {\rm ClientUpdate}(E_{L,i}^{t-1},  E_{T,i}^{t-1}, M_{i}^{t-1}, \tau, \lambda)
\end{equation}
After local training, the model will send only the language encoder $E_{L,i}^{t}$ to the server for aggregation as lines 9,11 in Alg.~\ref{pre explore alg}. In this way, the language encoder will be jointly updated on data from all the participated environments, thus being more generalized. Meanwhile, to further improve the generalizability of the language encoder, we randomly sample a fraction of seen environments at each communication round, where agents will also follow the training process above.   
The trajectory encoding module $E_{T,i}$ and multi-modal decision module $M_{i}$ will keep training locally, which can help local agents adapt to their own environments. For validation, we used the local models after local training. The whole training procedure is in Alg.~\ref{pre explore alg}.  

\begin{algorithm} [t]
    \caption{Federated Pre-exploration}
    \label{pre explore alg}
    \begin{algorithmic}[1]
        \STATE Parameters: Seen participation rate $r_{1}$, unseen participation rate $r_{2}$; local learning rate $\lambda$; server learning rate $\eta$; number of communication rounds $T$; number of seen environments $n$; number of unseen environments $m$; local training epochs $\tau$. 
        \STATE Initialize: $E_{L,i}^{0} = E_{L}^{0}$, $E_{T,i}^{0} = E_{T}^{0}$, $M_{i}^{0} = M^{0}$, for i in \{1,2,...,n+m\}
        \FOR{t in [1,T]}
            \STATE Server sample $r_{1}n$ seen environments  and $r_{2}m$ unseen environments as $\phi_{t}$\
            \STATE Server send global language encoder to selected environments $E^{t-1}$
            \FOR{client in $\phi_{t}$}
                \STATE Client update language encoder: $E_{i}^{t-1}=E^{t-1}$
                \STATE Client local training: $E_{L,i}^{t}, E_{T,i}^{t}, M_{i}^{t} = {\rm ClientUpdate}(E_{L,i}^{t-1},  E_{T,i}^{t-1}, M_{i}^{t-1}, \tau, \lambda)$
                \STATE Client upload delta of the language encoder $\Delta E_{i}^{t}=E_{i}^{t}-E^{t-1}$ to the server 
            \ENDFOR
            \STATE Server update language encoder: $E_{i}^{t} = E^{t-1} + \eta \sum_{i \in \phi_{t}} \frac{n_{j}}{\sum_{j\in \phi_{t}}n_{j}}  \Delta E_{i}^{t}$
        \ENDFOR
    \end{algorithmic}
\end{algorithm}

\section{Experimental Setup}

\subsection{Datasets}
We implement our federated learning framework on two datasets: Room-to-Room (R2R)~\cite{r2r} and Room-across-Room (RxR)(en)~\cite{rxr}. Both datasets are developed on the Matterport3D Simulator~\cite{r2r}, a photorealistic 3D environment for embodied AI research.

\noindent\textbf{R2R~\cite{r2r}} is constructed by generating the shortest paths from sampled start and end points. Then collect three associated navigation instructions for each path using Amazon Mechanical Turk (AMT). The dataset contains 7,189 paths from 90 environments, and each path contains 3 instructions. The environments are split into 61 environments for training and seen validation, 11 for unseen validation, and 18 for testing. The environments in unseen validation and unseen test set do not appear in the training environments. 

\noindent\textbf{RxR~\cite{rxr}} is proposed to mitigate shortcomings of former VLN datasets. Specifically, it is a large-scale dataset with multilingual instructions. It contains 16,522 paths and 126,069 instructions, among which 42,002 instructions are in English. RxR also ensures spatiotemporal alignments between instructions, visual percepts, and actions for agent training. The RxR dataset samples arbitrary paths from point to point (not necessarily shortest paths) to avoid data bias.

\subsection{Evaluation Metrics}
For both datasets, we report Success Rate (SR), Success Rate weighted by Path Length (SPL), Oracle Success Rate (OSR),  and navigation Error (NE) as goal-oriented metrics. SR is calculated as the percentage of the agent stop within 3 meters from the end point. SPL~\cite{spl} is defined as Success weighted by normalized inverse Path Length, which considers both navigation effectiveness and efficiency. OSR is the percentage of the agent visiting a point within 3 meters from the end point. NE is the average distance between the agent's final location and the end point. We also report Coverage weighted by Length Score (CLS)~\cite{jain-etal-2019-stay} and normalized Dynamic Time Warping (nDTW)~\cite{ndtw} to validate the fidelity of navigation paths, which penalize the deviation from the reference path. SR and SPL are often considered as the primary metrics for VLN evaluation. 

\subsection{Baselines}
Currently, we do not consider pre-training privacy, and VLN data pre-training infringes on data privacy. Thus, we choose two strong VLN baselines without VLN pre-training for experiments:
\begin{enumerate}
  \item \textbf{Envdrop}~\cite{Envdrop}: the environment dropout model uses a Bi-directional LSTM as the language encoder, an attentive LSTM as the action decoder, and a mixed learning objective of imitation learning and reinforcement learning. 
  \item \textbf{CLIP-ViL}~\cite{CLIP_on_vl}: the CLIP-ViL model adapts CLIP~\cite{clip} visual encoder to improve vision and language encoding and matching for vision-and-language navigation.
\end{enumerate}

\subsection{Implementation Details}

When training on seen environments, all models are trained till convergence. At each communication round, we use the participation rate of $r=0.2$, and train each local agent for $\tau=3$ epochs on local data for Envdrop model and $\tau=5$ for CLIP-ViL model.\footnote{The discussion and ablation study of local training epochs is in appendix.} For federated speaker training, we set the local epochs $\tau=5$ and select the best model on seen validation data according to BLEU score to generate instructions. 

During pre-exploration, we test the participation rate of $r_{1}=\{0.5,0.6,0.7\}$ for unseen environments. And we train each agent for $\tau_{1}=1$ epoch over unseen local dataset as the model converges quickly. When training across seen and unseen environments, we use the participation rate of $r_{2}=0.18$ for seen environments. To validate the effectiveness of our framework, we use federated trained speaker to generate pseudo-instruction and train from federated trained navigation agent for centralized pre-exploration, environment-based pre-exploration and federated pre-exploration.

\section{Results}
\subsection{Decentralized Federated Training} \label{results seen training}
\begin{table}[t]
\centering
\setlength{\abovecaptionskip}{8pt}
\setlength{\belowcaptionskip}{8pt}
\resizebox{\columnwidth}{!}{
\begin{tabular}{l|cccccc|cccccc}
\toprule
\multirow{2}*{\textbf{Model}} &
    \multicolumn{6}{c|}{\textbf{Val-Seen}} & \multicolumn{6}{c}{\textbf{Val-Unseen}}\\
    \cmidrule(lr){2-7}\cmidrule(lr){8-13}
    & NE$\downarrow$ & OSR$\uparrow$ & SPL$\uparrow$ & SR$\uparrow$ & CLS$\uparrow$ & nDTW$\uparrow$ & NE$\downarrow$ & OSR$\uparrow$ & SPL$\uparrow$ & SR$\uparrow$ & CLS$\uparrow$ & nDTW$\uparrow$ \\
    \midrule
Envdrop & 4.71 & 65.6 & 53.2 & 56.1 & 66.8 & 55.0
& 5.87 & 52.7 & 40.9 & 44.5 & 57.1 & 42.3 \\
FedEnvdrop &  5.20 & 60.2 & 48.3 & 51.2 & 64.3 & 51.2
& 5.52 & 55.5 & 43.9	& 47.5  & 59.2 & 45.7\\
\hdashline[1pt/2pt]
Envdrop$\rm _{aug}$ &  3.93 & 71.6 & 61.0 &	64.1 & 71.4 & 61.3
& 5.36 & 57.7 &	45.8 &	49.9  & 59.3 & 45.3  \\ 
FedEnvdrop$\rm _{aug}$ & 4.14 & 70.4 &	58.7 &	62.0 & 69.8 & 59.2 & 5.22 & 58.5	& 47.5	& 51.3 & 60.8 & 47.0  \\ 
\midrule

CLIP-ViL & 4.07 & 70.7	& 57.9	& 62.9 & 67.7 & 55.8
& 5.02 & 63.1 & 47.5	& 53.6  & 58.1 & 44.5\\
FedCLIP-ViL & 4.13 & 66.5 & 57.1 & 60.6 & 68.0 & 55.6
& 4.91 & 61.1	& 49.0	& 53.4 & 60.8 & 47.8 \\
\hdashline[1pt/2pt]
CLIP-ViL$\rm _{aug}$ & 3.52 & 75.0	& 61.7 & 66.8 &  69.3 & 58.6 
& 4.59 & 67.4	& 50.7	& 57.0 & 59.2 & 46.4 \\ 
FedCLIP-ViL$\rm _{aug}$ & 3.69 & 71.4 & 60.1 & 64.6 & 68.9 & 57.2 & 4.53 & 65.6 & 51.7 & 56.9 & 61.0 & 48.3\\ 
\bottomrule
\end{tabular}
}
\caption{R2R Results of seen environment training. Envdrop is the centralized Envdrop model, and FedEnvdrop is the federated Envdrop model. $\rm Envdrop_{aug}$ means the Envdrop model trained with augmented data. Our decentralized federated training outpeforms centralized training with Envdrop and achieves comparable results with CLIP-ViL on unseen environments.
}
\label{seen r2r}
\end{table}

\begin{table}[t]
\centering
\setlength{\abovecaptionskip}{8pt}
\setlength{\belowcaptionskip}{8pt}
\resizebox{\columnwidth}{!}{
\begin{tabular}{l|cccccc|cccccc}
\toprule
\multirow{2}*{\textbf{Model}} &
    \multicolumn{6}{c|}{\textbf{Val-Seen}} & \multicolumn{6}{c}{\textbf{Val-Unseen}}\\
    \cmidrule(lr){2-7}\cmidrule(lr){8-13}
    & NE$\downarrow$ & OSR$\uparrow$ & SPL$\uparrow$ & SR$\uparrow$ & CLS$\uparrow$ & nDTW$\uparrow$ & NE$\downarrow$ & OSR$\uparrow$ & SPL$\uparrow$ & SR$\uparrow$ & CLS$\uparrow$ & nDTW$\uparrow$ \\
    \midrule
Envdrop & 7.97 & 51.6 & 38.0 & 40.7 & 58.8 & 54.0 & 8.42 & 45.1 & 31.8 & 35.0 & 55.6 & 50.6 \\
FedEnvdrop & 8.30 & 53.3 & 35.9 & 39.7  & 57.2 & 51.8 & 8.58 & 47.4 & 30.2 & 34.4 & 55.4 & 48.8 \\ 
\midrule
CLIP-ViL & 6.92 & 56.8 & 42.3 &  46.5 & 60.0 & 56.2 & 7.38 & 50.6 & 34.9 & 39.5 & 55.6 & 51.2 \\
FedCLIP-ViL & 7.31 & 52.0 & 39.7 & 43.5 & 59.4 & 55.1 & 7.41 & 48.8 & 35.0 & 39.2 & 57.5 & 53.1 \\ 
\bottomrule
\end{tabular}
}
\caption{RxR results of seen environment training. Decentralized federated training obtains comparable results with centralized training on unseen environments (e.g., only 0.1\% SPL difference with the CLIP-ViL model).}
\label{seen rxr}
\end{table}

\noindent\textbf{Original data training} 
In Table~\ref{seen r2r} and Table~\ref{seen rxr}, we report the results for seen environment training on R2R and RxR datasets for both baselines. 

First, federated learning performs worse than centralized training on seen environments with an average of 2.8\% SR gap. This is reasonable, as centralized training can easily overfit to the seen training data for better performance on seen environments, while for federated learning, because of the decentralized optimization over protected local data, the global model can not overfit to the seen environments as well as centralized training. 

The performance on unseen environments tests the generalization ability of VLN models and is used for VLN evaluation.
As shown in Table~\ref{seen r2r} and Table~\ref{seen rxr}, on unseen environments, decentralized federated training achieves comparable results with centralized training on original data training across different VLN models. For example, FedEnvdrop performs better than Envdrop on R2R and nearly the same on RxR, and FedCLIP-ViL obtains comparable results with CLIP-ViL on both R2R and RxR. 
Thus, in terms of generalization ability, our decentralized federated training method is comparable with centralized training while protecting the training data privacy. 

\begin{table}[t]
\centering
\setlength{\abovecaptionskip}{8pt}
\setlength{\belowcaptionskip}{8pt}
\begin{tabular}{lcc}
        \toprule
        \textbf{Speaker} & \textbf{Val-seen} & \textbf{Val-unseen}  \\ 
        \midrule
        CLIP Cent  & 33.5 & 30.2  \\
        CLIP Fed  & 31.3 & 31.6  \\
        \hdashline[1pt/2pt]
        ResNet Cent & 33.6 & 30.7  \\
        ResNet Fed & 31.7 & 31.9  \\
        \bottomrule
    \end{tabular}
    \caption{Comparison of BLEU score between federated speaker and centralized speaker based on the CLIP encoder and the ResNet encoder on R2R.}
    \label{speaker bleu}
\end{table}

\noindent\textbf{Augmented data training} 
First, the performance of speaker model determines the quality of augmented data, and thus influences the performance of both augmented training and pre-exploration. Table~\ref{speaker bleu} shows that federated speaker performs 2.05 worse than centralized speaker on seen validation on BLEU score, but is 1.3 better on unseen validation data, which is quite aligned with the navigation results. And thus federated trained speaker is comparable with centralized speaker on instruction generation.

From Table~\ref{seen r2r}, when training on augmented data, the performance of federated learning is also comparable with centralized training on unseen environments, although the quality of pseudo data generated by federated speaker might be worse according to BLEU score on validation seen data. This shows that federated agent can learn generalized navigation strategies on pseudo data with lower quality.

\subsection{Federated Pre-exploration} 
To validate the effectiveness of federated pre-exploration on unseen environments, we compare centralized pre-exploration, environment-based pre-exploration, and different federated pre-exploration methods: full model sharing (Fed-Full), sharing language encoder only (Fed-Lan), and sharing language encoder across seen and unseen environments (Fed-Lan+seen). Results are shown in Table~\ref{pre-pxplore table}. 

\noindent\textbf{Navigation performance} For centralized pre-exploration and Fed-Full, in which one agent is optimized on data from all the environments, the agent can not adapt very well on each specific environment. For example, there is a gap of 3.1\% on SR between centralized training and environment-based pre-exploration. When sharing only the language encoder during federated learning, the validation results improve significantly comparing with full model sharing (e.g. 4.2\% on SR and 4.4\% on nDTW) since the agents can adapt to each environment better. Also, the generalization ability of language encoder is better than environment-based pre-exploration, since it is trained on more data across different environments. Thus, even under federated optimization, sharing only the language encoder in federated pre-exploration achieves similar results comparing with environment-based pre-exploration. 
Federated pre-exploration with seen environments further improves the performance benefiting from human labeled data, and achieves around 0.8\% SR improvement than environment-based pre-exploration. To sum up, our Fed-Lan+seen method achieves superior navigation performance in terms of both navigation success and path fidelity metrics. 

\noindent\textbf{Degree of privacy} From the perspective of privacy preserving, environment-based pre-exploration is the best, where nothing in the unseen environments will be shared with others. 
Centralized training is clearly the worse, where all the observable data from unseen environments will be directly shared with the server. Federated pre-exploration only uploads the model updates to the server.
Among federated methods, sharing only the language encoder protects data privacy better than full model sharing: it only shares the updates of language encoder, which accounts for only 24.6\% of the parameters and keeps other modules completely local.
Training with seen environments will not make the training process less private, as seen environments already shared their parameter updates with the server in decentralized federated training process. 

 \begin{table*}[t]
    \centering
    \setlength{\abovecaptionskip}{8pt}
\setlength{\belowcaptionskip}{8pt}
\resizebox{\columnwidth}{!}{
    \begin{tabular}{llccccccl}
        \toprule
        Model & Method & NE$\downarrow$ & OSR$\uparrow$ & SPL$\uparrow$ & SR$\uparrow$ & CLS$\uparrow$ & nDTW$\uparrow$ & \qquad \quad \; \;   Privacy$\uparrow$\\ \midrule
        \multirow{5}*{Envdrop} 
        & Centralized &3.89 &73.7 &61.7 &64.8 &71.5 &64.6 & 0 - sharing data\\
        & Env-based &\underline{3.49} &\textbf{78.5} &\underline{64.0}& \underline{67.4}&\underline{73.2} &\underline{67.5} & \textbf{3} - no sharing\\
        &Fed-Full & 3.96 & 74.1 & 59.1 & 62.4 & 70.9 & 63.3 & 1 - model sharing (100\%)\\
        &Fed-Lan &3.52 & 77.6 & 63.6 & 67.2 & \underline{73.2} & \textbf{67.6} &\underline{2} - model sharing (24.6\%)\\
        &Fed-Lan+seen &\textbf{3.47} &\underline{78.1} &\textbf{64.8} &\textbf{68.3} &\textbf{73.5} &67.3 & \underline{2} - model sharing (24.6\%)\\
        
        \midrule
        \multirow{5}*{CLIP-ViT} 
        & Centralized &3.70 &76.0 &60.8 &65.3 &70.5 &62.1 & 0 - sharing data\\
        & Env-based &\underline{3.31} &\underline{79.2} &65.2 &68.9 &\textbf{74.4} &\textbf{69.3} & \textbf{3} - no sharing\\
        &Fed-Full&3.68 & 74.9& 61.8& 65.8 & 70.5 & 61.9 & 1 - model sharing (100\%)\\
        &Fed-Lan &3.33 &\underline{79.2} &\textbf{65.4} &\underline{69.1} &74.0 &68.3 &\underline{2} - model sharing (24.6\%)\\
        &Fed-Lan+seen &\textbf{3.25} &\textbf{80.6} &\textbf{65.4} &\textbf{69.5} &\textbf{74.4} &\underline{68.9} &\underline{2} - model sharing (24.6\%)\\
        \bottomrule
    \end{tabular}
    }
    \caption{Comparison between different pre-exploration methods on R2R unseen validation. Fed-Full means full model sharing federated learning, Fed-Lan means sharing only language encoder in federated learning, Fed-Enc+seen means federated training with seen environments and sharing encoder only.}
    \label{pre-pxplore table}
\end{table*}

Overall, our federated pre-exploration method achieves a good performance-privacy trade-off. Centralized training is both worst in terms of navigation ability and privacy protection. Environment-based pre-exploration has the best privacy protection of unseen environment data. Federated pre-exploration achieves the best navigation results with little privacy cost by keeping all client data locally, and sharing only the language encoder model updates with the server.

\section{Conclusion and Future Work}

In this paper, we study the data privacy problems in vision-and-language navigation with respect to two learning scenarios: seen environment training and unseen environment pre-exploration. We propose a novel federated vision-and-language navigation (FedVLN) framework to preserve data privacy in two learning stages while maintaining comparable navigation performance. Furthermore, we present that federated pre-exploration can even outperforms all previous pre-exploration methods and achieves the best performance-privacy trade-off.
As the first work along this direction, our work does not consider adversarial attacks that can potentially recover data information from shared local model updates, and we believe future work can consider more embodied AI tasks and defend against privacy attacks for more data security.
\\

\noindent\textbf{Acknowledgement}
We thank Jing Gu, Eliana Stefani, Winson Chen, Yang Liu, Hao Tan, Pengchuan Zhang, and anonymous reviewers for their valuable feedback.
This work is partially supported by the PI's UCSC start-up funding.

\clearpage
%
%
\bibliographystyle{splncs04}
\bibliography{cameraReadyNew}
\clearpage
\appendix

\section{Case analysis}

\begin{figure}
\centering
\subfigure[Environment-based]{
\begin{minipage}[t]{0.45\textwidth}
\centering
\includegraphics[width=5.5cm]{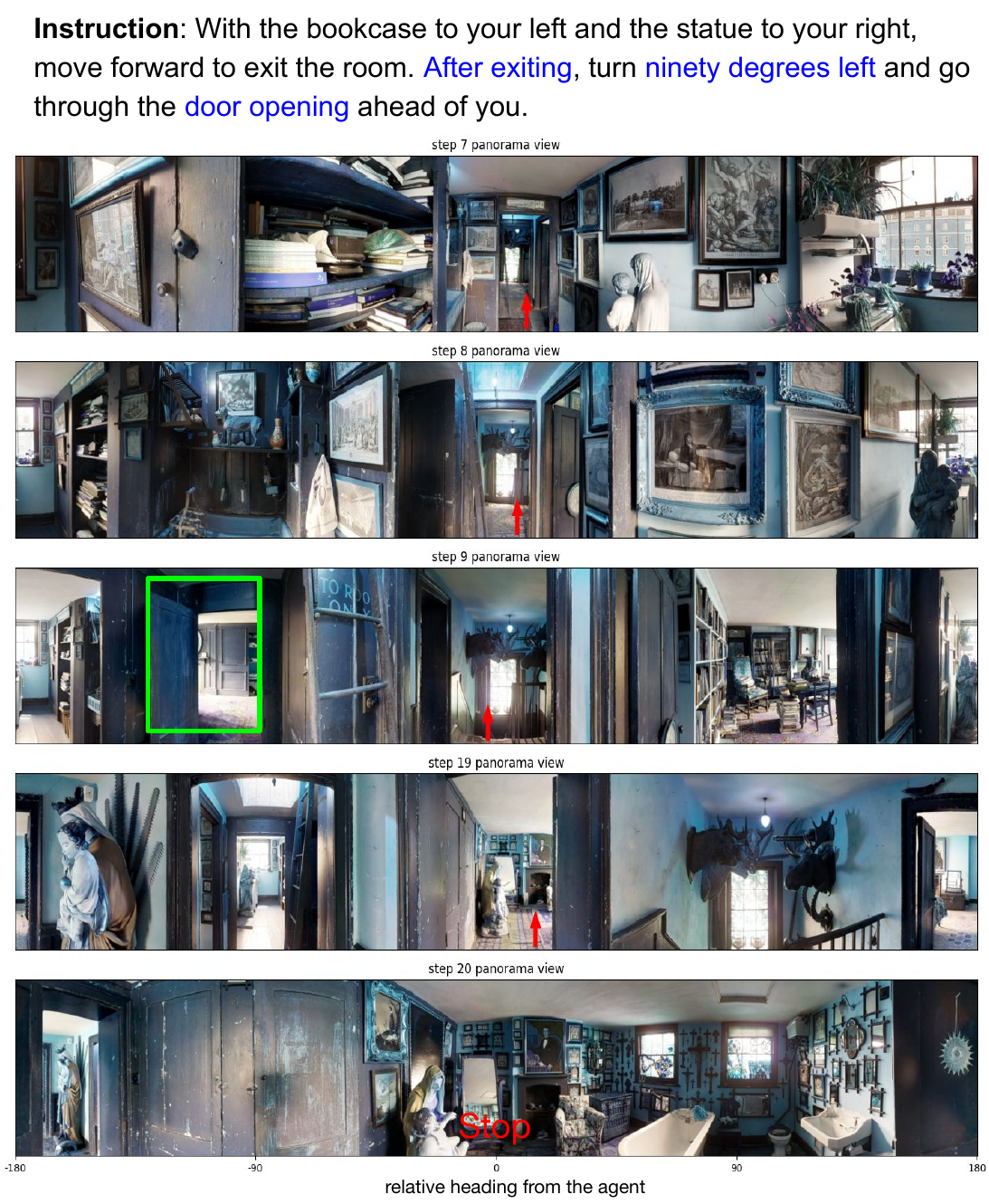}
\end{minipage}
\label{fig:env fail}
}
\subfigure[Fed-Enc]{
\begin{minipage}[t]{0.45\textwidth}
\centering
\includegraphics[width=5.5cm]{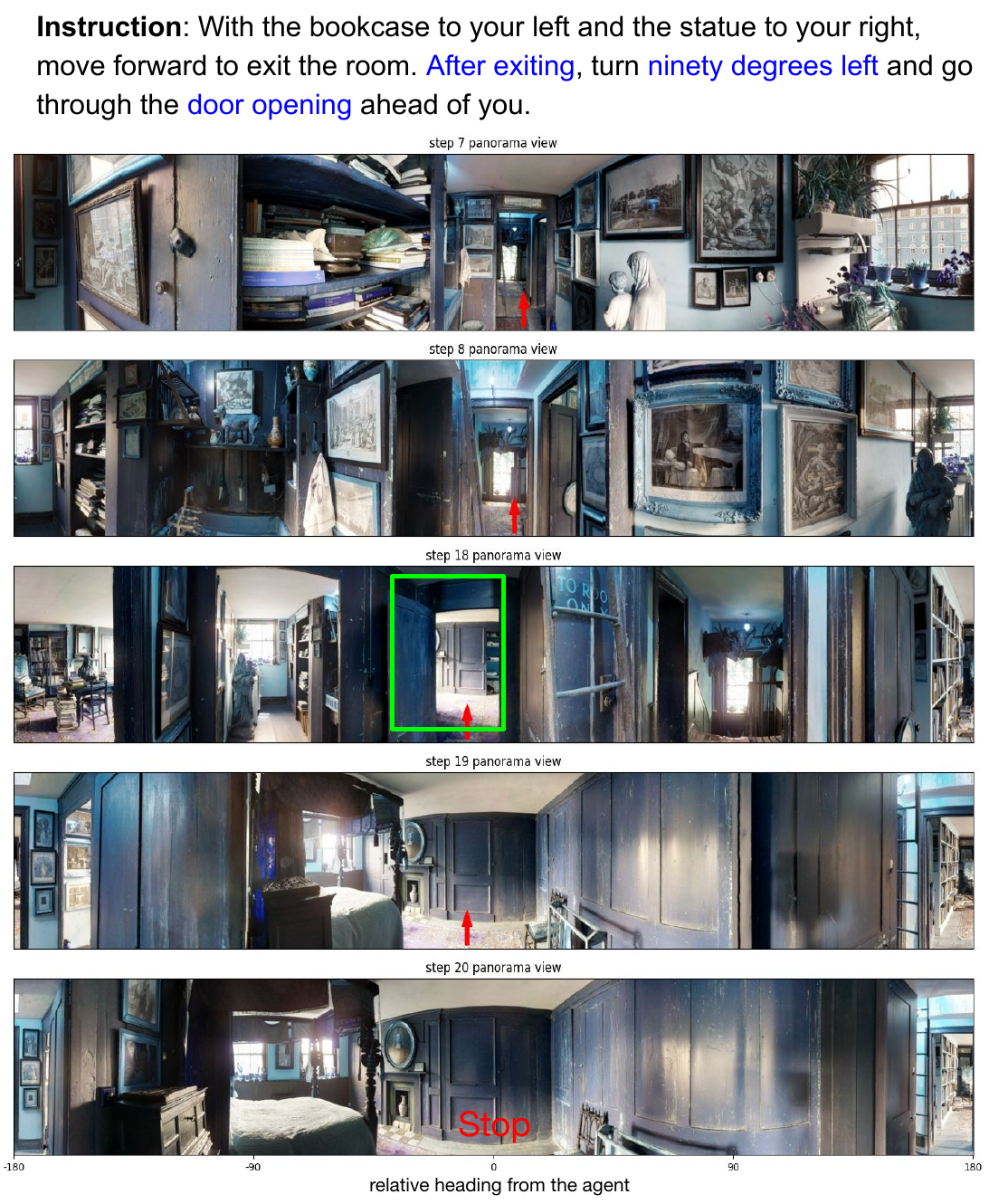}
\end{minipage}
\label{fig:unseen success}
}
\caption{Case study between environment-based pre-exploration and language encoder-sharing federated pre-exploration. In this example, the agent trained by environment-based pre-exploration can not fully understand that there will be an open door on the left as soon as it exits the room.} 
\label{env vs unseen}
\end{figure}

In this section, we further demonstrate the advantage of sharing language encoder and training with seen environments during federated pre-exploration by visualizing the navigation trajectory examples from environment-based pre-exploration~\cite{APS}, language encoder-sharing federated pre-exploration and federated pre-exploration sharing language encoder across seen and unseen environments. We also show how centralized training overfit better than federated training on seen environment. 

First, the agent trained by environment-based pre-exploration can not understand the language instruction and execute it precisely. Fig~\ref{env vs unseen} demonstrates a qualitative example for this. The environment-based pre-explored agent successfully exits the room, but it does not notice the opened door at 90 degrees left and keeps going ahead, and finally enters the wrong room. The agent trained by sharing language encoder in federated pre-exploration can understand the words `after exiting', `ninety degrees left', and `door opening' precisely and notice the open door right after it exits the room, thus navigating successfully. 

\begin{figure}[t]
\centering
\subfigure[Fed-Enc]{
\begin{minipage}[t]{0.45\textwidth}
\centering
\includegraphics[width=5.5cm]{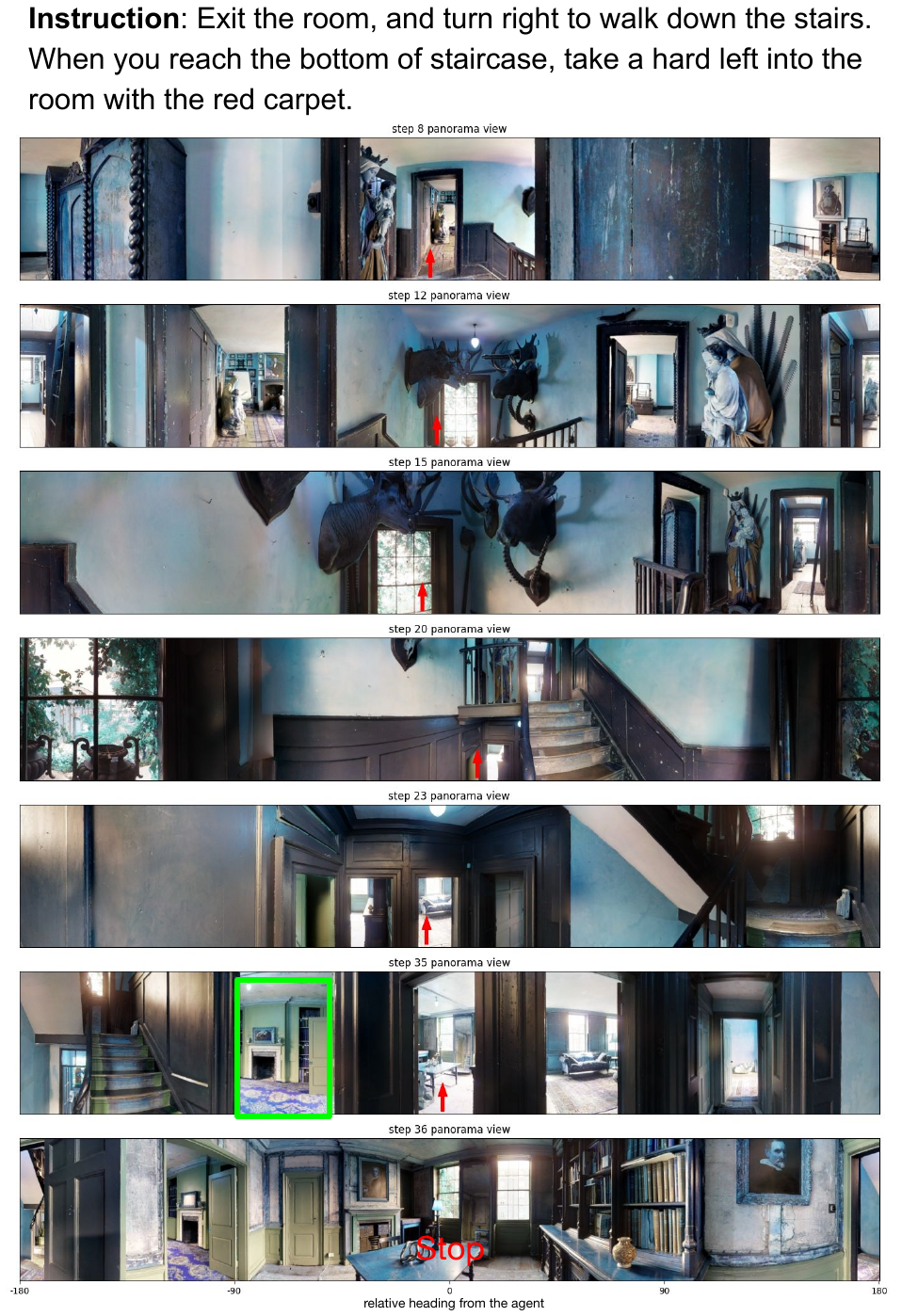}
\end{minipage}
\label{fig:sampleFed-Enc}
}
\subfigure[Fed-Enc+seen]{
\begin{minipage}[t]{0.45\textwidth}
\centering
\includegraphics[width=5.5cm]{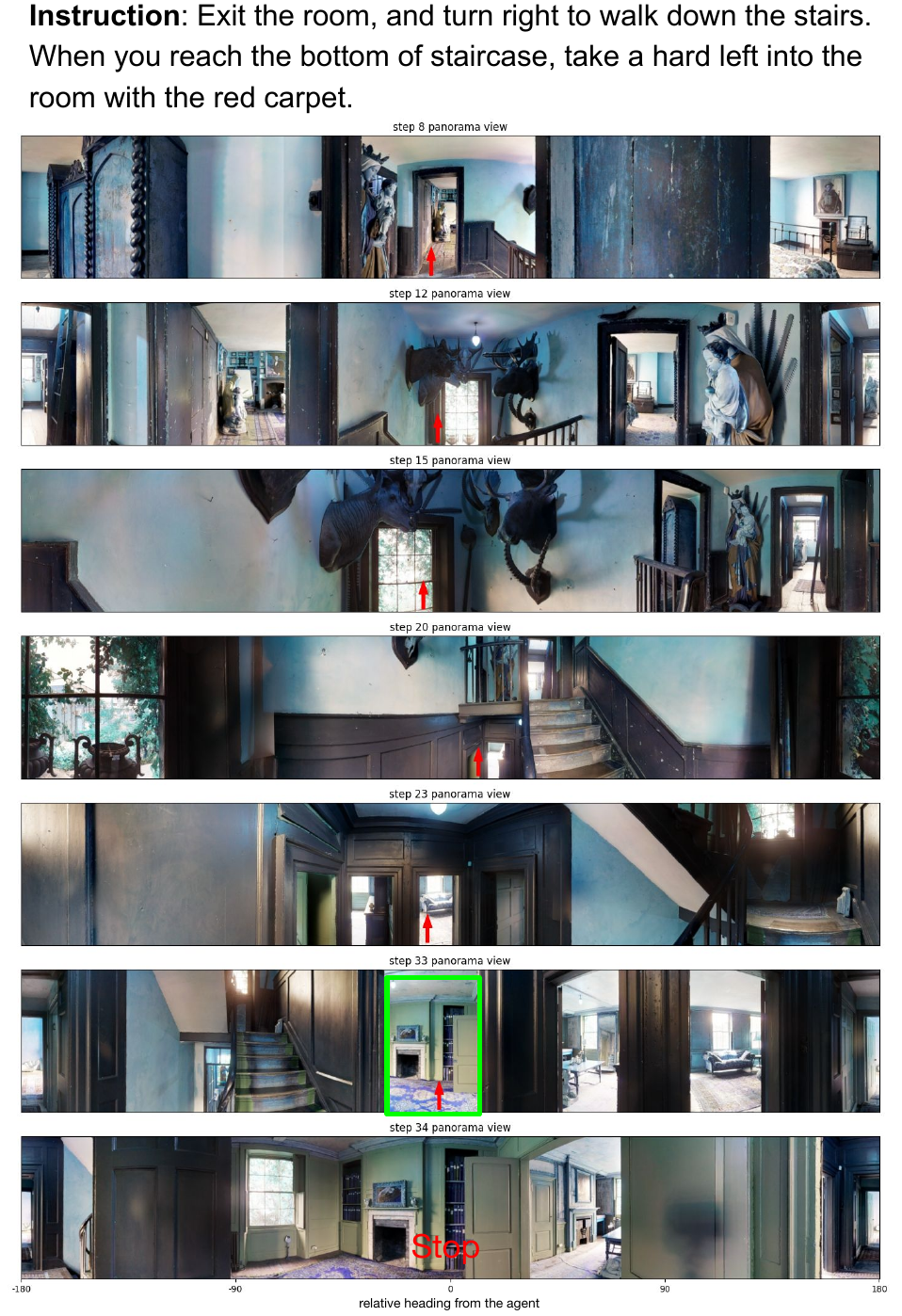}
\end{minipage}
\label{fig:sampleFed-Enc+seen}
}
\caption{Case study between language encoder-sharing federated pre-exploration with and without seen environments. In this example, the agent trained by federated pre-exploration without seen environments can not understand and notice the keyword `red carpet'.} 
\label{fig:seen vs unseen}
\end{figure}

Second, federated pre-exploration training with seen environments can further improve the language understanding ability of the agent, especially for words that the speaker generates rarely. In Fig~\ref{fig:seen vs unseen}, the agent trained by federated pre-exploration sharing encoder without seen environments navigates successfully at most part of the path. However, it can not understand `red carpet' well and target the correct direction, thus entering the wrong room at the end, although the red carpet is within its vision all the time. The agent trained by federated pre-exploration with seen environments recognizes the red carpet and navigates successfully.  

\begin{figure}[t]
\centering
\subfigure[Federated training]{
\begin{minipage}[t]{0.47\textwidth}
\centering
\includegraphics[width=5.7cm]{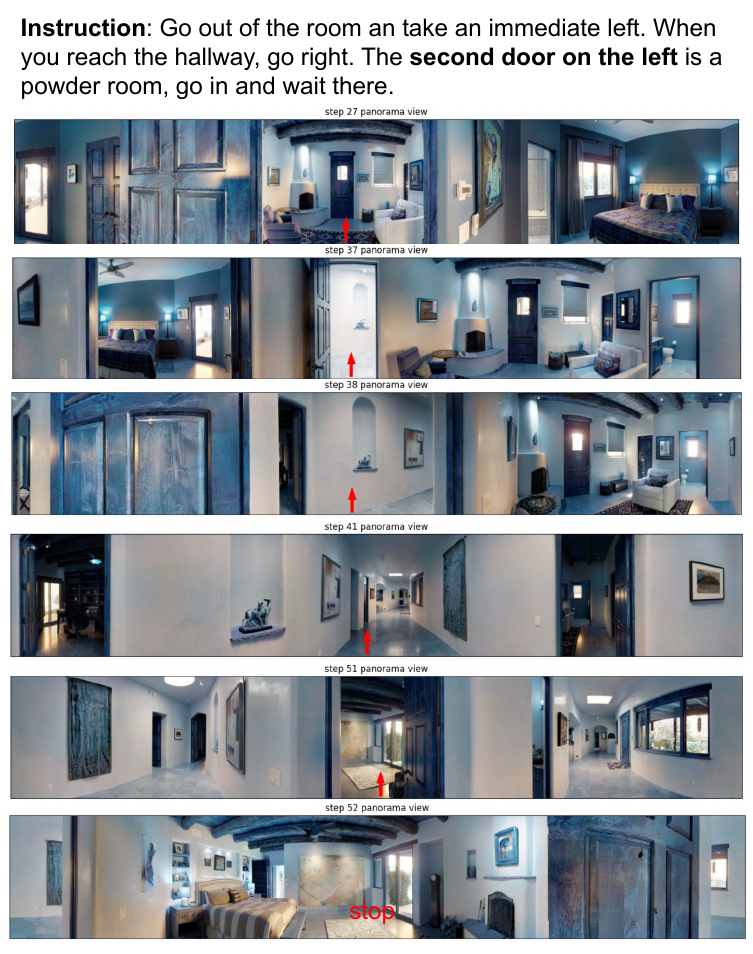}
\end{minipage}
\label{fig:samplefed}
}
\subfigure[Centralized training]{
\begin{minipage}[t]{0.47\textwidth}
\centering
\includegraphics[width=5.7cm]{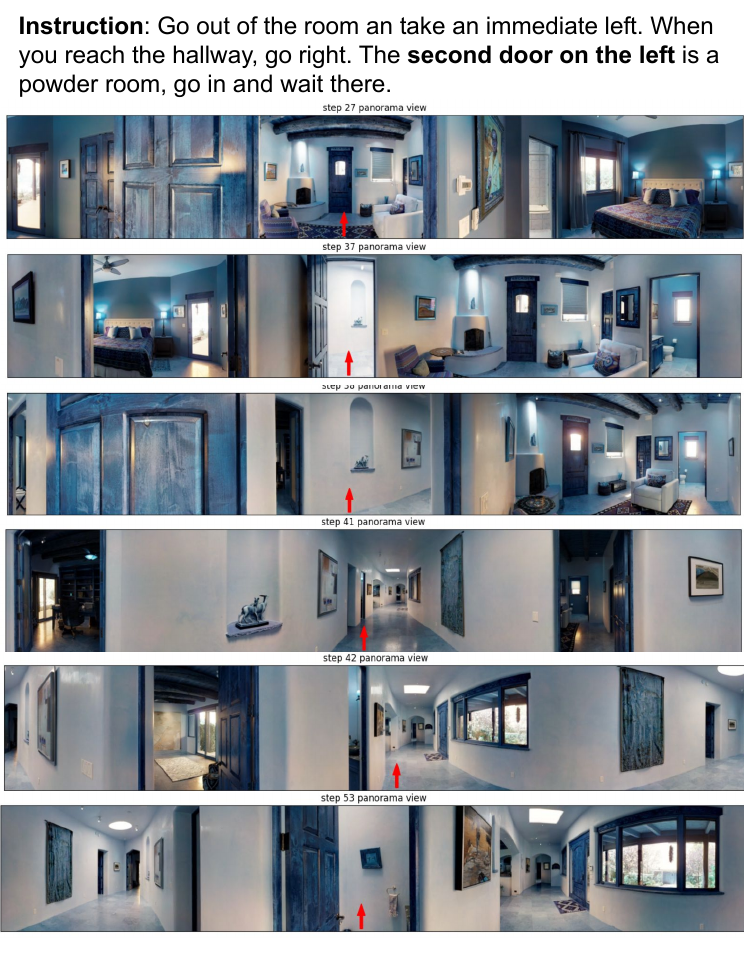}
\end{minipage}
\label{fig:samplecent}
}
\caption{Case study between decentralized federated training and centralized training on seen environment. } 
\label{fig:cent vs fed seen}
\end{figure}

 \begin{table*}[hbt!]
    \centering
    \setlength{\abovecaptionskip}{8pt}
    \setlength{\belowcaptionskip}{8pt}
    \begin{tabular}{cc}
        \toprule
        training & valid seen \\
        \midrule
        \makecell{ ``f4939bf6f00a4864832a358f1ea8394e", \\
      ``28c09c307b11487c999f88e1e9ec3231", \\
      ``ecff9ecf0cfe4d8bb83260fc092f3b00", \\
      ``d1ffe5280fce4ac5a949cdc9ee8b6f7c", \\
      ``dbc0fd77d9384e14a6d0a19302b85a15", \\
      ``4ff62efbc0934e888120522e4c84e712", \\
      ``982920829a0b433880410222539f240e" } 
        & 
        \makecell{``fae83673fc694cd9a18c215ce6d92c58", \\
         ``28c09c307b11487c999f88e1e9ec3231", \\
         ``ecff9ecf0cfe4d8bb83260fc092f3b00", \\
         ``d1ffe5280fce4ac5a949cdc9ee8b6f7c", \\
         ``dbc0fd77d9384e14a6d0a19302b85a15", \\
         ``4ff62efbc0934e888120522e4c84e712", \\
         ``982920829a0b433880410222539f240e" }\\
        \bottomrule
    \end{tabular}
    
    \caption{Comparison of two similar paths in training set and validation seen. The path in validation seen is used in Fig.~\ref{fig:cent vs fed seen}.}
    \label{tab:training vs seen}
\end{table*}

Third, the agent trained by centralized training memorized path appeared in the training data better than federated training. From Table.~\ref{tab:training vs seen}, we can see that the chosen example validation path is highly overlap with a training path with the same later part. In this example, as shown in Fig.~\ref{fig:cent vs fed seen}, centralized training successfully navigate to the location, while federated trained agent fail at the later part of the path and choose the wrong room in the hallway. 

\section{Discussion of performance-convergence speed trade-off}

In federated learning, the convergence speed is also an important indicator to evaluate the model except for model performance. In federated vision-and-language navigation, faster convergence speed means fewer communication rounds towards a target success rate.  The local models in the environments can thus achieve satisfactory navigation performance sooner. Also, fewer communication rounds lead to less communication overhead~\cite{fedavg} and privacy leakage. Thus, in this section, we discuss the influence of a key hyper-parameters, communication frequency on convergence speed and model performance by ablation studies. 

\begin{table*}[t]
    \centering
    \setlength{\abovecaptionskip}{8pt}
\setlength{\belowcaptionskip}{8pt}
\setlength\tabcolsep{5pt}
\begin{minipage}{0.45\textwidth} 
\centering
\begin{tabular}{lccc|c}
        \toprule
        E \quad & 43 & 45  & 47& Best\\ \midrule
         3 &234 &400 &480 & \textbf{48.2} \\
         5 &183 &291 &847 & 47.0 \\
         8 &112 &204 & -- & 46.0 \\
        12 & 79 &\textbf{115} & -- & 46.3 \\
        18 & \textbf{65} & 152 & -- & 45.6 \\
        \bottomrule
    \end{tabular}
    \caption{Comparison of communication rounds needed to achieve target success rates(\%) on R2R unseen validation between different local epochs (E) based on Envdrop model. The last column is best unseen success rate.}
    \label{tab:le-com envdrop}
\end{minipage}
\begin{minipage}{0.45\textwidth} 
\centering
    \begin{tabular}{lccc|c}
        \toprule
        E \quad & 48 & 50 & 52& Best\\ \midrule
         3 &461 &665 &1151 & 52.4 \\
         5 &266 &367 &570& \textbf{53.4} \\
        8 &219 &318 &514& 52.4 \\
        12 & 129 &209 & -- & 51.8 \\
        18 &\textbf{101} &\underline{181} & -- & 51.2 \\
        \bottomrule
    \end{tabular}
    \caption{Comparison of communication rounds needed to achieve target success rates(\%) on R2R unseen validation between different local epochs (E) based on CLIP-ViT model. The last column is best unseen success rate.
    }
    \label{tab:le-com clip}
\end{minipage}
\end{table*}

As shown in Table~\ref{tab:le-com envdrop} and Table~\ref{tab:le-com clip}, overall, using small local epochs can achieve better performance at the end. However, the communication rounds it takes may be unacceptable and impractical in application. Using larger local epochs leads to faster convergence, while it suffers from local over-fitting and can not achieve better navigation performance. Thus, it's better to set local epochs to a middle size: 5-12. 

\section{Comparison of training speed}

\begin{table}[t]
    \centering
    \setlength{\abovecaptionskip}{8pt}
    \setlength{\belowcaptionskip}{8pt}
    \begin{tabular}{lcc|cccc}
        \toprule
        Metrics & Cent & Fed &  Cent & Env &  Fed-Full & FedLan+seen  \\ \midrule
        Step (${\rm 10^{2}}$) & 356 & 825 & 105 & 7 & 97 & 13 \\
        \bottomrule
    \end{tabular}
    \caption{We evaluate the training steps that the model needs to achieve a certain unseen success rate. The first two columns are for seen environment training without augmentation towards success rate of 44\% for Envdrop and 52\% for CLIP-ViL, last four columns are for pre-exploration towards success rate of 62\% for Envdrop and 65\% for CLIP-ViL.} 
    \label{tab:training time}
\end{table}

We here evaluate the training cost and efficiency of FedVLN. On seen environment training, the average training steps towards certain SR are $3.56\times 10^{4}$ for centralized training, and $8.25\times 10^{4}$ for federated training. However, since federated learning can utilize the computation power of edge devices, the training step can not fully represent training time comparison between centralized training and federated learning. 

On pre-exploration, our final federated pre-exploration method uses only $1.30\times 10^{3}$ steps to achieve certain performance, which improves over full model sharing federated learning and centralized training. The training steps of `FedLan+seen' is larger than environment-based pre-exploration but not by a large margin. 
More importantly, our federated learning framework does not bring any extra cost during inference and maintains high inference efficiency.

\end{document}